# Responses to a Critique of Artificial Moral Agents

**Adam Poulsen\***
Charles Sturt University,
Bathurst NSW, Australia
apouls02@postoffice.csu.edu.au

**Michael Anderson**
University in Hartford,
Connecticut, United States
anderson@hartford.edu

**Susan Leigh Anderson**
University of Connecticut,
Connecticut, United States
susan.anderson@uconn.edu

**Ben Byford**
Machine Ethics Podcast &
Ethical by Design, Bristol,
United Kingdom
hello@benbyford.com

**Fabio Fossa**
Sant'Anna School of Advanced
Studies, Pisa, Italy
fabio.fossa@santannapisa.it

**Erica L. Neely**
Ohio Northern University, Ohio,
United States
e-neely@onu.edu

**Alejandro Rosas**
Universidad Nacional de
Colombia, Bogotá, Colombia
arosasl@unal.edu.co

**Alan Winfield**
University of the West of
England, Bristol, United
Kingdom
alan.winfield@uwe.ac.uk

**Abstract**

The field of machine ethics is concerned with the question of how to embed ethical behaviors, or a means to determine ethical behaviors, into artificial intelligence (AI) systems. The goal is to produce artificial moral agents (AMAs) that are either implicitly ethical (designed to avoid unethical consequences) or explicitly ethical (designed to behave ethically). Van Wynsberghe and Robbins' (2018) paper *Critiquing the Reasons for Making Artificial Moral Agents* critically addresses the reasons offered by machine ethicists for pursuing AMA research; this paper, co-authored by machine ethicists and commentators, aims to contribute to the machine ethics conversation by responding to that critique. The reasons for developing AMAs discussed in van Wynsberghe and Robbins (2018) are: it is inevitable that they will be developed; the prevention of harm; the necessity for public trust; the prevention of immoral use; such machines are better moral reasoners than humans, and building these machines would lead to a better understanding of human morality. In this paper, each co-author addresses those reasons in turn. In so doing, this paper demonstrates that the reasons critiqued are not shared by all co-authors; each machine ethicist has their own reasons for researching AMAs. But while we express a diverse range of views on each of the six reasons in van Wynsberghe and Robbins' critique, we nevertheless share the opinion that the scientific study of AMAs has considerable value.

# Introduction

The field of AI ethics is broadly divided into two main branches. By far the largest of these branches is concerned with the ethical *application* of robots and AIs; the important question of how human developers should behave in order to minimize the ethical harms that can arise from robots and AIs in society, either because of poor (unethical) design, inappropriate application, or misuse. This branch is generally referred to as either AI or robot ethics and has already led to the development of ethical principles (Boden et al. 2017; Boddington 2017), standards (International Organization for Standardization 2016), and good practice (IEEE Standards Association 2017).

The smaller branch of AI ethics is concerned with the question of how robots and AIs can *themselves* behave ethically. Referred to as ethical AI, ethical robots or – more generally – *machine ethics*, this field of study is both philosophical (addressing questions such as should a machine be ethical and, if so, which ethics should it embody?) and technical – addressing the considerable scientific and engineering challenges of how to realize an ethical machine. The philosophical foundations of machine ethics where set out by authors participating in the 2005 AAAI Fall Symposium (https://www.aaai.org/Library/Symposia/Fall/fs05-06.php). Wallach and Allen (2009) coined the term 'artificial moral agent' (AMA) that we use here.

This paper is concerned with machine ethics (while mindful of the wider context of AI ethics). Our primary aim is to respond to van Wynsberghe and Robbins (2018) paper *Critiquing the Reasons for Making Artificial Moral Agents*. The 'reasons' for developing AMAs discussed in that paper are: it is inevitable that they will be developed; the prevention of harm; the necessity for public trust; the prevention of immoral use; such machines are better moral reasoners than humans, and building these machines would lead to a better understanding of human morality. In this paper, each co-author addresses each of those six reasons in turn.

Machine ethics research is still in its infancy. It follows that machine ethicists are not unified in their research approach and goals, or on the wisdom of the real-world application of AMAs. Simply, machine ethicists discuss most aspects of AMAs agreeing only that pursuing the research is useful and worthwhile. This is to be expected in all fields of ethics. The discussions in machine ethics are comparable to those in general ethics. In general ethics, some ethicists favor utilitarianism while others favor Rossian deontology, and so on. In machine ethics, some machine ethicists favor a certain type of ethical agency over another, one domain of application over another, and so on. Furthermore, machine ethicists discuss whether AMAs should be used at all. Some think that all machines should be AMAs while others would disagree. Just as general ethicists, machine ethicists discuss the details while endeavoring to define what ought to be ethical for moral machines. To demonstrate the diversity of the machine ethics approaches, consider AMA ethical agency. The different types of ethical agency considered are defined by Moor (2006). Moor categorizes the degree of the ethical agency an AMA possesses, while also considering the implications of those AMAs. For each machine ethicist, their ultimate intentions for AMA use falls within one of the agency types. Machine ethicists often advocate different types of ethical agency, which is to say they think AMAs should – within Moor's schema – be either implicit, explicit, or full ethical agents[1].

Machine ethicists also discuss both which domains AMAs should be used in, and when they should be used. The domain might be elder care, industrial robotics, social robotics, etc. The question of use is concerned with when, or in what situations, it is appropriate for a robot to be an AMA. A crossover occurs when a machine ethicist believes that all robots should be AMAs, and thus deployed in all domains. Not all machine ethicists would agree, instead some might say that there are cases when a robot should not be an AMA depending on a myriad of factors such as the complexity of the human-robot interaction (HRI), the vulnerability of the user, the usefulness of a robot being an AMA in a certain situation, and so on.

Evidently, since there are many types of AMAs and discussed aspects of AMAs, each machine ethicist has their own research approach and goals, and opinion on AMA use. The same goes for each author, or author team, of this paper.

*Adam Poulsen*
*Where it is possible and sensible, robots should be explicit AMAs. Instances where it is not possible is when there is a lack of artificial intelligence present, a robot ought to be sufficiently intelligent to make good, informed moral decisions. Cases where it is not sensible is where a user, or any other stakeholder, experiences physical or psychological harm caused by the AMA because of it making dynamic ethical decisions which overlap safety measures or its core goals. Although I advocate for explicit AMAs, in the absence of absolute moral truths AMAs should only have limited moral freedom. That is, they should only be allowed to make ethical decisions within predetermined, limited situations. I apply this concept to the design of care robots where I suggest that such machines should be built be intrinsically safe and thus limited. However, they should also have the freedom to make explicit ethical decisions to customize care to individual patients based on user values.*

*Susan Leigh Anderson & Michael Anderson*
*We are developing and deploying principles and practices that ensure the ethical behavior of autonomous systems to the end of advancing beneficial AI development as well as the study of ethics.*

*Ben Byford*
*As introduced by van Wynsberghe and Robbins' (2018) in their critique of AMA research it would seem that while "autonomy in robots and AI increases so does the likelihood that they encounter situations which are morally salient". It seems to follow that if this is to continue and autonomous systems like automated vehicles and robotic care workers are to be relied upon for their desired purpose we should indeed endeavor to research and experiment within a burgeoning field of enquiry,*

---

[1] Moor's ethical impact agents have not been included here because such agents have no moral decision-making capabilities (the focus of machine ethics).

including: Machine Ethics, Robot Ethics, Business ethics, Data ethics and Design Ethics. These endeavors and how they coalesce, as mentioned in this paper's introduction, are still in process and not yet ready to make categorical recommendations to industry. AMAs are but a part of a larger puzzle caused by the desire to create AI systems with greater levels of capacity and autonomy. My hope and endeavor is to produce work that will enable us collectively to create more positive autonomous systems for society.

*Fabio Fossa*
In my research I mainly focus on problems arising from the way in which "artificial agents" are conceived of. I think that it is necessary to highlight the risk of using ideas and words that normally apply to human beings in order to describe forms of machine functioning–particularly since, at least to an extent, this may be inevitable. Anthropomorphism is of course the most critical issue in this domain, and still a very troubling one. In my opinion, it is of great importance to increase awareness of the influence that the use of language exercises on the way problems are framed and tackled. More specifically, I think that the debate concerning AMAs is in many cases hindered by the concepts and words that are chosen to present the issues at hand. I believe that discussing the way in which concepts as "agency" and "morality" change when applied to machines may help create adequate expectations towards machine ethics and identify the value of potential AMAs.

*Erica L. Neely*
I am interested in ethical issues surrounding emerging technologies such as AMAs and artificial intelligence. I believe it prudent to consider the ethical issues that may arise from these technologies before the technology is fully deployed. Thus, while we are some distance from creating a true artificial intelligence, say, it would behoove us not to wait until that point to grapple with ethical questions raised by that technology. We have already seen that ethical issues arise from autonomous vehicles – these issues will only become more pressing as we develop even more sophisticated machines.

*Alejandro Rosas*
In my contribution I attempt to clarify why artificial agents will require capacities for autonomous moral reasoning and decision making. As AI reaches human level intelligence, i.e., the ability to have knowledge in any arbitrary domain of facts, including social facts and social relationship to other agents, AI will reach the status of a person and will need moral knowledge and decision-making capacities. In artificial agents, however, moral freedom should never be absolute: they should not be allowed the freedom to jump outside of moral thinking for self-interested reasons, as humans do when they act against moral norms that are socially or legally sanctioned.

*Alan Winfield*
This author regards research in AMAs as a scientific endeavor; one that seeks to answer the question: is it in principle possible to computationally model moral behavior? This approach is less concerned with the utility of AMAs (and remains skeptical of the wisdom of real world AMAs), than in the scientific question of what *cognitive machinery is required to realize an explicitly ethical machine and how such a machine performs in experimental trials. By computationally modeling simple ethical behaviors it is hoped that we might gain new perspectives and insights into what it means to be a moral agent. This author is convinced that ethical behavior requires inter alia an agent to be able to model and internally represent itself and other agents and to predict (anticipate) the consequences of both its and others' actions (Winfield and Hafner 2018).*

It is clear that each machine ethicist has their own approach and vision for AMA research. Within those several contexts, the following sections showcase the voices of each author separately in addressing the critique of reasons for developing AMAs in van Wynsberghe and Robbins (2018), section by section. Since the purpose of this paper is to add to the conversation, the authors either defend or disagree with the reasons for developing AMAs. The reasons critiqued by van Wynsberghe and Robbins for developing AMAs, which will be addressed in order, are: it is inevitable that they will be developed, the prevention of harm, the necessity for public trust, the prevention of immoral use, such machines are better moral reasoners than humans, and building these machines would lead to a better understanding of human morality. Finally, after each reason has been discussed, there will be an overall conclusion.

## Discussion of the Reasons for Developing Moral Machines

### Inevitability

*Adam Poulsen*
I do not agree with the notion that explicit, or full, ethical agents (from here on in referred to as AMAs in my contribution) should be developed because they are inevitable. It is simply my position that the development of AMAs might prove to be useful as autonomous ethical helpers in situations where there is a lack of humans, especially in the healthcare sector. It is not that AMAs will inevitability be created, it is that they will inevitability become necessary in healthcare at least due to the global shortage of caregivers (Burmeister, 2016; Burmeister, Bernoth, Dietsch, & Cleary, 2016; Burmeister & Kreps, 2018; Draper & Sorell, 2017; Garner, Powell, & Carr, 2016; Landau, 2013; Sharkey & Sharkey, 2012a, 2012b; Sparrow & Sparrow, 2006; Tokunaga, Tamamizu, Saiki, Nakamura, & Yasuda, 2017; Vallor, 2011). The following are direct responses to statements from van Wynsberghe and Robbins' (2018) article.

For the clarity requested by van Wynsberghe and Robbins (2018) concerning when to make use of AMAs, consider the following. The time and place for when and where a robot should be an AMA are up for discussion in machine ethics. My position is that AMAs should not be used in every situation, only in ones where it is sensible - such as healthcare – and only when limited moral freedom is enforced. An AMA that has limited moral freedom is one which can only make morally reasoned dynamic actions within predetermined scenarios i.e. its dynamic decision making is restricted by the setting and its end goals. As stated in my introduction, for it to be sensible to use an AMA, a user and other stakeholders must not experience physical or psychological harm caused by the AMA because of it making dynamic ethical decisions which overlap safety measures or its core goals. To justify this position, I will explore the usefulness of AMAs in robot-delivered care.

Healthcare is a context where interactions should be customized based on the subjective interaction with the care recipient to ensure that they have the best experience possible with personalized care. That is, good care is determinative in practice (Beauchamp 2004); decisions made in situ within the caregiver-care recipient relationship are key to good care (Upton 2011). Thus, care robots should be customizable and capable of making dynamic decisions in run-time, as I have argued previously (Poulsen and Burmeister 2019; Poulsen et al. 2018a; Poulsen et al. 2018b; Poulsen et al. forthcoming). Care robots, where it is sensible, should be AMAs with ethical reasoning to customize behaviors and interactions to ensure better quality, individualized care. The notion that it must be sensible to make a care robot an AMA makes it improbable for a surgical robot to sensibly be an AMA. The reason for this is that a patient might be physically harmed if a surgical robot's core goals are disrupted by it making ethical decisions which change the HRI. Some robots do not have to participate in HRIs that require gathering personal data about the user and making person-driven dynamic actions for an optimal or improved ethical operation, a surgical robot is an example of this. Such robots far from personal HRIs, all that needs to be considered is safety and this can be ensured by good design or implicit ethical agency. Thus, AMAs should not be used in every situation, they should only be used where the customization of actions is sensible, such as healthcare.

Current care robots are simply "delegated a moral role" (van Wynsberghe and Robbins 2018) as a therapy dog is, but not all care robots need be. A care robot that is not an AMA is functional, however, it is not concerned with an individual user's values. If it could be, and if the jobs it undertakes deems ethical decision-making use sensible, then it should be an AMA. If it was possible to make a therapy dog in an elderly care facility a "moral dog" (van Wynsberghe and Robbins 2018) (one with ethical reasoning), would that not be a better experience for the users? Yes, as outlandish as it is, if one could have a therapy dog understand their values and have it making ethical decisions to determine its actions, then that would make for a better experience. The same could be said for a care robot. Consider the damage a robot may cause if it is not considerate of a user's values, the interpretation of those values, and the user's value priorities. For example, what if an elderly care robot, that is simply 'delegated a moral role' and is not an AMA, cannot reason that an elderly person needs social company and thus their value of social connectedness suffers. Today's care robots are not AMAs, but they should be in the sense described here.

### *Susan Leigh Anderson & Michael Anderson*
van Wynsberghe and Robbins (2018) claim that a distinction must be made between "being in a morally charged situation" and "being delegated a moral role". Fair enough. They cite the example of Corti that advises a human on the risk of a caller on the phone having a heart attack, with the human making the final decision, as one where the machine satisfies the first, but not the second condition. Then the authors claim: "If one agrees with machine ethicists then one should accept that it is inevitable that a moral role reserved for the human in this case will be assigned to the machine" (van Wynsberghe and Robbins 2018). There are three things wrong with this statement and the general thrust of this section countering the need for machine ethics on the grounds that its being needed is inevitable:

1. We maintain that there may be cases where we do not want machines to make ethical decisions and act on them, recommending keeping humans in the loop, when the ethics is not agreed upon, for example.
2. The "inevitability" that we have foreseen is the fact that *autonomously functioning machines are being developed and their actions are ethically significant.* (Incidentally, this does not necessarily involve "life and death".) So, our position is that there is a need for machine ethics for *these machines* is inevitable.
3. Concerning the authors use of the term "moral role" as applied to robots, there is ambiguity. For us, an ethical machine is capable of making and acting on ethical decisions, but not as humans are thought to be able to do so. They lack the qualities that enable humans to be held morally responsible for their actions, so are not full ethical agents. We think this is a good feature because we do not want these machines to be able to act unethically and we ultimately want to put the blame for unethical behavior of the machine on its human designers and builders.

**Ben Byford**
Van Wynsberghe and Robbins' (2018) Internet of Things (IoT) argument of becoming "simply untenable" seems to point out the obvious while deliberately leaving out any technical deliberation. As both a researcher and technologist one is often faced with both the philosophical and the practical. In the cited case I agree with the argument that "harm" should be categorized to include other forms of indirect harm, not simply physical. This seems appropriate in the age of digital banking, social media, etc. What may be an oversight is that AMAs could be added to IoT networks, if needed, supporting any designed safety rules, as these types of networks can (most do) connect directly with a hub node. This node could be in the shape of a central server run by an organization, a computer run locally to the network, or a series of programs run on a distributed network protocol using blockchain technology. As a few simple examples it is plain to me that AMAs could be applicably used for IoT and indeed in other similar networks.

Concerning van Wynsberghe and Robbins (2018) Corti example, the system works well as an illustration of human-machine collaboration within a specific domain (in this case heart attack signal diagnosis). I applaud the use of technology in this way and hope that more organizations can harness AI to benefit or enhance their current functions. However, in many domains it is impractical to wait for human decision-making mostly as the amount of data and speed of decision-making render it impractical for humans to react fast enough; we simply would not be able to make a decision that was in anyway considered. The obvious example here would be autonomous vehicles; however, we could also talk about stock market algorithms, drones, and other robots in a similar way. For autonomous vehicles, research suggests that during a ride "to expect humans to remain alert may be overly optimistic" (Goodall 2014). This means that for these types of vehicles it may prove safer to never defer control to the occupant, or indeed never with a short reaction time. As Fabio Fossa points out we would do well to allow AMA systems, or at least continued research thereof, within this domain as any other options seems plainly suboptimal.

It seems to me the ship has already sailed on restricting AIs or autonomous algorithms solely to non-ethical decision-making contexts and it is now in societies best interest that researchers are able to create new ways to make these technologies safer, more predictable, and event more intelligent – I believe this includes research in AMA by extension.

### Fabio Fossa
I am not fully convinced by van Wynsberghe and Robbins' (2018) claim against the technological necessity of AMAs. The starting point of their reasoning is, in my opinion, too broad and must therefore be further specified. If the necessity of AMAs is at stake, we are thus discussing about autonomous technologies, i.e., artefacts that can execute specified functions without necessarily requiring human intervention or supervision. When autonomous technologies are involved, the delegation of moral roles to machines is not a matter of specific choice. On the contrary, such delegation depends on the general features of the automatized task.

If we decide to automatize a task which does not require in any sense the exercise of moral agency when carried out by human beings, then the need for moral machines do not arise. On the contrary, if a task does require some form of moral agency when carried out by human beings, then delegating the same task to autonomous machines necessarily implies delegating a moral role as well.

Sure enough, we can choose to strictly forbid the delegation of such tasks to autonomous machines. It is true that there is always a human decision at the outset concerning whether or not to delegate a task to autonomous machines. However, I think it would be naïve to expect such decisions to be taken with purely moral considerations in mind.

We may then address this issue by programming machines to freeze when they face situations that calls for moral evaluation. Such machines, however, would not function autonomously, and their appeal would drop. Thus, we might try to build machines capable of selecting autonomously between alternatives in accordance with moral constraints. Such machines would not require human intervention, while at the same time carrying out their tasks following some agreed moral values – like Poulsen and Burmeister's (2019) dynamic value trade-offs in run-time – provided we can identify these values, translate them into computer language, and finally build machines that function in accordance to them, which is of course far from trivial and perhaps impossible.

Be it as it may; however, if autonomous machines will be deployed to carry out tasks which, when carried out by human beings, display a moral side, then either we choose not to address this side, or we must find a way to implement some form of morally relevant data elaboration and action selection into the machines themselves.

This, of course, would not make artificial moral agency equal to human moral agency at all. It would instead provide us with autonomous tools that would not offend our moral sensitivity and would satisfy our moral expectations.

Consider for instance the case of autonomous cars. The point of autonomous cars is to free human beings from driving themselves, a task that arguably requires moral agency when carried out. This purpose would not be accomplished if autonomous cars required human intervention to deal with driving situations. Moreover, the computing speed of such technologies extends well beyond the reach of our responsiveness, so that real-time human intervention may not be an option (as Wiener (2013), 118 already pointed out). This being the case, we can (a) forbid the technology to be researched and hit the market, (b) choose not to address the moral side of the 'driving' task and to develop autonomous cars with zero moral sensitivity (which, of course, does not mean that they would be unsafe, since moral constraints are other than safety constraints), or (c) we need to build morally sensitive autonomous cars, i.e., morally sensitive autonomous executers of the 'driving' function. To me, (c) seems to be the better option, thus worthy of being explored in practice.

Moreover, in my opinion the 'Corti example' fails to prove the authors' point and, by the same token, underlines the confusion that surrounds the notion of AMA here under examination. Corti, being a "support system" (van Wynsberghe and Robbins 2018), is not an AMA. In fact, it is not an autonomous agent: it requires human intervention to accomplish the purpose it is built to serve, i.e., to send help as soon as possible to people who show signs of suffering a heart attack while calling the emergency services. Corti offers support that may result in a more efficient execution of a task which is, however, substantially carried out by human beings. Therefore, Corti is not an AMA; it would be an AMA had we built a fully automated emergency service system and delegated all the tasks carried out by human phone operators to it. If, for some reason, we would want to do this, then I think that the system should be designed in a way that, for instance, human phone operators at the emergency serviced would find morally satisfactory, to the extent that their task implies the exercise of moral agency. AMAs enter the picture only when a task is fully automated, which requires the exercise of moral agency when carried out by human beings. This "moral" side of autonomous machine functioning can only be properly addressed by design.

### Erica L. Neely
As Susan Leigh Anderson & Michael Anderson acknowledge, autonomously functioning machines are already being developed. Humans are very good at creating things first and only later considering whether we should have done so; as such, we have already begun to create autonomous machines, even without settling questions about machine ethics. Similarly, there is a great deal of curiosity about the mind and the nature of consciousness which, in part, drives some members of the AI community. This curiosity is likely to propel us to the point that we eventually develop true AI, regardless of whether it is ultimately a good idea.

But what happens when we do? The authors' example is of a (limited) artificial intelligence in a very specific, constrained circumstance: Corti evaluates emergency phone calls and determines the likelihood that the caller is having a heart attack; it then provides that information to a human who ultimately makes the decision on how to act. The authors place weight on the idea that being an AMA requires having a machine be "delegated a moral role" (van Wynsberghe and Robbins 2018) and they question whether this is inevitable in any sense. In the case of Corti, the answer is probably no; Corti has a fairly simple task and is unlikely to require being an AMA to accomplish that task (although it is not clear that being an AMA would hinder the process either). There is no need to delegate a moral role in this circumstance.

However, not every robot or AI is designed for a single specific purpose. Some of them are intended to be artificial general intelligence (AGI). AGIs function broadly and, as such, they are going to end up in a variety of situations; some of these will be morally salient and some not – just like for humans. This sort of autonomous machine will likely need to be an AMA in order to navigate the spectrum of interactions it will have. It would be neither practical nor perhaps even desirable to have a human intervening in every interaction; the idea of an autonomous machine, after all, is that it can function on its own.

While one might argue that we would be better off not designing this kind of general AI, I believe that is essentially too late. Machines exist along a spectrum. At one end we have toasters and microwaves, which lack interests and rationality and exist to fulfill a very narrow purpose; concepts such as moral agency and patiency simply do not apply to them. At the other end, we have true AIs, which I believe would have interests and the rationality to evaluate moral situations; these could be both moral agents and patients. We are not to that point yet. Currently, we are somewhere in the middle, which is where the complication lies. For any given instance, it is difficult to know whether it would have to be an AMA. But in denying inevitability, the authors are claiming that we will hit a hard stop on this spectrum – and that is implausible to me. We have already started down that path, and it seems to be human nature to keep pushing the limits of knowledge (and creating things simply because we can).

*Alejandro Rosas*
The question about the inevitability of AMAs is crucial. I use the term "AMA" to refer only to robots that are explicit ethical agents (those that have an explicit set of normative principles, which they can utilize during decision-making). At some point in the future the scope of the epistemic competence of artificial agents will reach domain-universality. AI will necessarily include in their knowledge social facts and institutions, which in turn include moral facts, i.e., mainly facts about the bads and goods that arise from the interaction between persons. It is also probable that such AI will be implemented in robots with the ability to move freely in the world: moving freely may even turn out to be a necessary condition for learning about the world.

The robots we are able to produce today are not like this. All the ethical capabilities they need can be programmed implicitly as in healthcare robots, for example. Their autonomous computations are restricted to establish the facts of the situation. The needs of a patient are just facts, which can be general, e.g. a permanent or temporary condition of the patient, or particular ones to be picked up from perceptual cues, including the patient's verbal requests. A healthcare robot can ethically respond to those cues without any moral reasoning. The robot follows implicit ethical rules, without any explicit practical reasoning using both facts and moral principles as premises. Similar to an automatic teller machine, it is designed to do the right thing, following such rules as: "if the facts are p, do x; if the facts are p', do x*, if the facts are p'', do x**," and so forth. The rules connecting facts to actions follow from ethical principles that are not represented in the robot's program. The robot does not use such principles, only the programmers do.

Implicit ethical agents of this sort can be said to lack moral freedom completely. Such artificial agents are designed to operate and reason in a limited domain of facts (e.g. monetary transactions, healthcare) with a limited set of applicable moral rules (honesty in monetary transactions, satisfaction of the needs of the sick and the elderly). In such cases it is always possible to design the robot ethically without an explicit ethical reasoning program. However, if the care robot is required to make the patient feel cared for in a sense related to love and friendship, this hardly seems possible without mimicking emotions and fine tuning them to a diversity of situations. Plausibly, this might require autonomous moral reasoning in a limited domain, to achieve a reasonable compromise between the requirements of healthcare goals and respect for the feelings of patients.

The inevitability of AMAs possessing moral freedom will arise with the expansion of their knowledge and reasoning capabilities to all domains of facts. If the time is reached when AI can autonomously and reliably learn and reason about facts in any domain, if this AI is implemented in robots and the needs of the labor market require them to move around freely, they will need to evaluate the harms that could affect both them and others. When this point is reached, they will inevitably require autonomous ethical decision-making as well, which of course will be used not only to avoid harming humans, but also to prevent humans from being harmed, i.e., to help humans (or persons generally).

Regarding the issue of responsibility, we need clarity about what sort of "unethical" behavior is possible for an intelligent machine, in contrast to a human. Will explicit ethical agents be able to perform unethical acts out of self-conceit? This depends on how they are programmed. They need not be programmed to trump a moral norm out of self-interest. If, however, we mean by "unethical" in a machine a controversial ethical decision in a moral dilemma, this is a very different meaning of "unethical", for it does not imply that the action violates rights out of self-interest. This use just reflects the fact that when rights – or moral views – conflict, as in moral dilemmas, any decision is likely to be labelled "unethical" by parties in disagreement about how to resolve the conflict. In these cases (for example, in "trolley-type" dilemmas faced by autonomous cars) the programming of the robot must be constrained by what the law forbids or demands. Such agents should most likely be designed as implicit ethical agents so as to comply with the law, with no need to reason themselves, beyond the reasoning needed to ascertain the facts of the situation. Failures of reasoning about facts leading to "unethical behavior" are obviously the responsibility of the designer.

*Alan Winfield*
I agree with van Wynsberghe and Robbins: AMAs are not inevitable. Here I am referring to *explicit* ethical agents; implicitly ethical agents are not only inevitable but a good thing. There are, at present, no explicit ethical agents in real-world use. The only explicit ethical agents that exist are a handful of proof-of-concept laboratory prototypes. These are minimally ethical machines with limited functionality designed only to test hypotheses about how to build such machines. None are examples of real-world robots – such as autonomous cars – with added ethics functionality. van Wynsberghe and Robbins' plea that "the development of commercially available AMAs should not proceed further" seems premature given that their commercial development has not started at all.

No one would question the view that *all* real-world robots must be developed and operated responsibly, and subject to tough standards and regulatory frameworks. If (and I do not believe this is inevitable) AMAs became a practical proposition such machines would, in my view, first need to be the subject of public consultation, then – in operation – subject to even greater levels of regulatory oversight, transparency, and ethical governance than regular robots.

**Artificial Moral Machines to Prevent Harm to Humans**

*Adam Poulsen*
Safe design, or implicit ethical agency, should be the priority. Designing an autonomous system to be an AMA with explicit ethical decision-making should be the second option and it ought to only be considered if it is sensible. Furthermore, the addition of explicitly ethical machine-made decisions into a system which is already implicitly ethical by safe design should not supersede implicit, safe-by-design actions. With this simple ordering of priorities, one can design an ethical machine which is both safe-by-design and capable of ethical decision-making at the same time. A machine does not necessarily have to be either implicitly ethical by safe design or explicitly ethical by the capacity to make dynamic decisions, it can be both with the former consideration always superseding the latter (Poulsen et al. 2018b). For example, an autonomous wheelchair could be designed to avoid hitting people by way of proximity sensors that detect objects ahead and force it to stop when something is near – this is a safe design. However, there is room for ethical decision-making in the case of this care robot. The autonomous wheelchair could also have the capacity to allow an impatient care recipient to take control of the wheelchair movement and speed, so they can and move around with more freedom – the ethical decision being made here is allowing the user to have more independence. But, if at any point the user is about to hit something or someone, safety will take priority and the wheelchair will lock the user out of the controls and it will return to an autonomous mode. In this example, the autonomous wheelchair need not be only safe nor only capable of ethical-decision making, it can be both with safety as the priority. A complete description of this approach is described elsewhere (Poulsen and Burmeister 2019).

How does this approach, or ordering of priorities, make AMAs better at preventing harm to humans? My position is that AMAs, such as the autonomous wheelchair described above, could help to prevent psychological harm, as well as physical harm since the former can erupt into the latter. This safety-first, Values in Motion Design approach (Poulsen et al. 2018a; 2018b) ensures that the physical safety of users is implicit in the robot's design. After intrinsic safety is guaranteed, any customized, user-centric considerations will be considered. Acknowledging user preferences with personalized care robot decisions would help to better the experience of persons within the HRI and improve the psychological condition of a user, especially in the case of care which is shown to be good if it is person-centered (McMillan et al. 2013; Santana et al. 2018). I have shown elsewhere in an empirical study made up of 102 participants, that a care robot which aims to respect an individual user's values and customize its actions according to those values is acceptable, usable, and value sensitive (Poulsen and Burmeister 2019). My contention is that a more ethically considerate, value sensitive robot might reduce psychological harm and ultimately physical harm. Moreover, this value sensitivity can only be achieved with an AMA capable of making dynamic decisions in run-time.

Van Wynsberghe and Robbins (2018) make several comments that require a direct response, they are as follows.

> "[Machine] ethicists may also agree with the pursuit of safe robots and then the real concern for ethicists is that ethics is being reduced to safety... if AMAs are simply a solution to possibly harmful machines, then *safety*—not *moral agency*—is the object of debate... machine ethicists must either distinguish what makes their machines "moral" above and beyond "safe" (van Wynsberghe and Robbins 2018).

AMA research is not just the pursuit of safe robots as the authors describe. Moreover, ethics cannot be reduced to safety. The general goal of AMA research is more complex, researchers want to create machines with autonomous ethical decision-making. The reasons critiqued by the authors are not the goals of AMA research itself, they are some of the reasons offered by researchers attempting to justify their pursuits. Thus, the reason critiqued here ('artificial moral machines to prevent harm to humans') is not a goal or pursuit of AMA research, it is a reason to continue it.

An intrinsically safe-by-design robot falls into what machine ethicists call implicit ethical agents (Moor 2006). Moor defines an implicit ethical agent as a machine that "acts ethically because its internal functions implicitly promote ethical behavior—or at least avoid unethical behavior" (2006). The reason for creating such a robot is to make it implicitly ethical or safe-by-design, without it having the capacity to make explicit ethical decisions. AMA research is primarily focused on developing explicit ethical agents, those with autonomous ethical decision-making (Moor 2006). This is not to say that AMAs should not be safe or be integrated into a safe-by-design machine with the Values in Motion Design approach described above. The important distinction being made here is that AMAs make explicit ethical decisions. Generally, AMA research attempts to implement central ethical considerations, such as principles and values, into the machine so it may use them to make explicitly ethical, real-time decisions. A machine which is safe-by-design is not generally a machine ethics pursuit because such a machine's actions are implicitly ethical and preprogrammed, without any decisions being made in situ. My position, however, is that these considerations can meet halfway within the Values in Motion Design approach demonstrated above with the autonomous wheelchair example.

### *Susan Leigh Anderson & Michael Anderson*
What do van Wynsberghe and Robbins (2018) mean by an "artificial moral machine"? What is supposed to be 'artificial'? If 'morality', is the morality that a machine follows supposed to be different from what should be imposed on humans? If so, we disagree. We, also, disagree with the scholars referred to in the lead sentence: "For many scholars the development of moral machines is aimed at preventing a robot from hurting human beings" (van Wynsberghe and Robbins 2018). Ethics is not only about preventing harm, but doing the *best* action in a given situation. This may even involve harming a human (e.g. a robot that is charged with caring for a young child might correctly push a child down, thus causing him some harm, to prevent him from running into the street as a speeding car is approaching).

Concerning using AMAs to prevent harm to humans, van Wynsberghe and Robbins argue that either AMA research is only concerned about possible *harm* to humans, in which case safety concerns can be addressed in the construction of machines or recommendations for their use; or machine ethicists desire something more than that, they want a true *moral* agent that *cares* for us and has *feelings* oneself, in which case only a human being will do. We have made four points in our work that are relevant here:
1. We only require that the machines behave in an ethically responsible manner, which includes doing more than preventing harm.
2. There is, unfortunately, a shortage of human beings to perform needed tasks (Draper and Sorell 2017; Sharkey and Sharkey 2012b, 2012a; Sparrow and Sparrow 2006; Tokunaga et al. 2017; Garner et al. 2016; Landau 2013; Vallor 2011; Burmeister 2016). Furthermore, it is ethically preferable for us to develop machines that can do these tasks rather than not, as long as we can ensure that they do so in an ethical manner. A good example is an eldercare assistant robot to do all sorts of things for a person who wants to remain at home when no trained human being is available to fulfill this role.
3. We maintain that having feelings oneself as being critical to behaving ethically is questionable. Human beings are prone to getting carried away by their feelings to the point where they behave unethically. It is better to follow an ethical principle than be subject to the whims of emotion, so we do not think ethical robots should have emotions. But, to behave ethically, it is essential that one appreciate the feelings of those who do have them, so ethical robots must be able to recognize the feelings of the humans it interacts with and deal with them properly.
4. We believe that it is wrong, ethically, to deliberately try to fool vulnerable

human beings (such as the very young, elderly, and physically or mentally challenged) into thinking that the robots that interact with them have feelings for them. Thus, the thrust of some work in robot development, along these lines, causes us some concern.

### Ben Byford

Within van Wynsberghe and Robbins' (2018) article there is often reference to safety as a preference to a human centric view of morals and ethical thinking while discussing machine autonomy. There may indeed be a linguistic "trojan horse" (van Wynsberghe and Robbins 2018) if we a conflating moral reasoning to a solely human pursuit.

In a useful thought experiment here: Alan Turing proposed that a machine that acts (or, famously in his experiment The Imitation Game, communicates with text) in a similar way to humans can be both confused for human. That is, there can be no meaningful way of distinguishing between the two without access to prior knowledge. In a similar way it can be said that morals or ethical decision-making can be exhibited by a machine, however we may not claim that the machine is being humanly moral as it may hold no *understanding* of morality. However, as with The Imitation Game, we come to the same outcome either way. I am happy to argue either way with someone saying machine algorithms can or cannot be moral but if we are only concerned with what is exhibited then it makes little difference to argue. As part of a system's safety features, it may be *safer* to add AMA capabilities. Whether you believe it is an ethical feature or safety one, or indeed whether the decision-making is humanly moralistic decision or not, it may produce the same outcome. In my view this argument is over semantics.

Secondly, allowing a system to only have safety features, which implies it was designed to have implicit ethical behavior only, we come to issues of inaction when faced with an ambiguous situation or a situation simply not accounted for within the system's coding. I believe an AMAs may be able to increase a system's autonomy in these cases and herald more positive interactions instead of none.

I slightly disagree with my co-author Alan Winfield's point regarding intention: the way an AMA is implemented may give rise to an impossibility to interrogate the system's intention. Though there is lots of research currently being done in this area, machine learning algorithms are mostly seen as black boxes. An AMA created with machine learning techniques would surely suffer from this whether the designer's moralistic intentions are good or not.

### Fabio Fossa

I believe that van Wynsberghe and Robbins' (2018) discussion of the relation between moral issues and safety issues arising from autonomous machine functioning is somehow confused. On page 7 of their article van Wynsberghe and Robbins (2018) correctly state that moral issues cannot be reduced to safety issues. Thus, it seems that moral issues arising from autonomous machine functioning must be specifically addressed. This is precisely what machine ethicists try to do: to build machines that function not only effectively, not only safely, but also in a morally satisfactory fashion - i.e., in a way that, ideally, would prevent moral harm and concur to the affirmation of moral goods. From this acknowledgement should follow, therefore, that safe robots are not, by the same token, moral machines. And yet, on page 14 van Wynsberghe and Robbins (2018) write: "One should not refer to moral machines, artificial moral agents, or ethical agents if the goal is really to create safe, reliable machines. Rather, they should be called what they are: safe robots". Why is that?

I agree that we should avoid any reference to the notion of agency when discussing morally sensitive autonomous executers of functions, since words as 'agency' and 'agent' too easily engender anthropomorphism and false expectations (I express similar concerns in Fossa (2017)). Yet, safety issues and moral issues are different from one another and only partially overlap, so that the label 'safe robots' would not meet the need that causes the introduction of labels such as 'AMAs' or 'moral machines'. Even though the techniques which may lead to safe *and* morally satisfactory machines may be analogous, safe machine functioning and moral machine functioning differ from one another. Imagine a future, say, one hundred years from now where a subject is suffering from a health condition that deeply weakens her but does not affect her judgment. Suppose that she is taken care of by a robot caregiver whose function (among others) is to feed and hydrate her safely – i.e., without physically hurting her. Now, suppose also that the patient decides not to eat anymore and let go. However, the robot keeps executing its functions. Even though no safety issues would arise, it may be argued that a moral issue is at stake. Clearly, safe robots can execute functions in a way that may be perceived as immoral by the human beings involved just as a hypothetical 'moral' robot may execute its functions in unsafe ways. Safe robots are not necessarily moral machines; and I do not see how to talk about this side of autonomous machine functioning if not as 'moral'–provided the essential difference between functional autonomous morality and full moral agency is borne in mind.

### Erica L. Neely

I agree that ethics is about more than simply preventing harm to humans. In their section on 'Machine Ethics' van Wynsberghe and Robbins reference Asimov's Three Laws of Robotics, which begin with the idea that a robot must never cause harm to a human; I regard this as unethical in the case of a true AI, since it would deny the AI a right to self-defense. Thus, I do not believe that we will necessarily seek to prevent harm in absolutely every case. I also think it would likely be impossible to do so; humans are very creative at finding new ways to cause harm. However, preventing harm in general (or in most cases) is a reasonable goal for an ethical agent – it just is not the only goal they would have. It could thus be part of an AMA's system of morality, even if it was not sufficient by itself. Similarly, the desire to make machines less likely to harm humans could be part of a justification for having AMAs, even if I am sympathetic to the idea that it is not sufficient justification on its own.

With that said, it is not at all clear to me why the "concept of *moral* machines or artificial *moral* agents invites, or more strongly requests that, the user believe the robot may care for him/her, or that the robot can experience feelings" (van

Wynsberghe and Robbins 2018emphasis in the original). Ethical theories are built upon specific principles. It is true that these principles can involve feelings (such as in an ethics of care) but usually they do not. Utilitarianism involves maximizing utility across a population. Kantian deontological ethics requires us to act in a fashion that is rationally universalisable. Virtue ethics requires us to cultivate particular character traits. None of these three theories are specifically linked to being able to feel.[2] Indeed, Kantian ethics seems particularly well-suited for AI, since it is supposed to apply to all rational beings and depends solely on rationality; it seems entirely possible that an AI could be a Kantian ethical agent, regardless of whether they have the ability to feel. While we should presumably not mislead people as to whether machines can care or feel, I do not think that ability is inextricably wedded to whether they can be AMAs.

*Alejandro Rosas*
Van Wynsberghe and Robbins (2018) state that it is unnecessary to endow a machine with autonomous ethical capabilities in order to prevent it from harming human beings. This is certainly correct. Many of the machines we use have some potential to harm us. To reduce this potential is to design them for safety. This can be done without endowing them with moral reasoning. But, sometimes robots are designed to help humans. And again, the plausibility of designing such robots as implicitly instead of explicitly ethical depends on the scope of their domain of application. An automatic teller machine helps humans get their money faster and easier. Because of its limited domain of application, it can be designed as an implicit ethical agent. Another example of a helping robot is Corti. The difference between an automatic teller machine and Corti is not primarily about ethical capabilities. Neither of them has those capabilities. They differ in their reliability concerning factual reasoning. An automatic teller machine is completely reliable concerning number crunching and can decide on its own, given the user's input, how much money to hand over to the user and how much to subtract from her account. Corti, in contrast, is not a completely reliable judge concerning the (factual) question whether someone on the other end of the phone-line is in danger of a heart attack. In order to be an implicit ethical agent like the teller machine, Corti must first reach human-level reasoning about facts in its restricted domain. Only then will it be allowed to make on its own a strictly limited decision to help, e.g., sending an ambulance to the person at the other end of the line. Why will it be unnecessary to endow it with ethical reasoning capabilities once it has reached this point? As in the case of the automatic teller machine, it is because of the limited context and the manageable set of moral rules needed, which can be accounted for implicitly.

The necessity of endowing robots with autonomous ethical capabilities will arise only with the expansion of their reasoning and knowledge capabilities – and of their freedom of movement and action – to a multiplicity of factual domains.

---

[2] Even in the case of utilitarianism, which is often seen as involving pain or pleasure, I have argued that the ability to feel is not necessary; any being with interests can be harmed or benefitted, thus utilitarian calculations can be performed. See (Neely 2013) for a further discussion of this topic, albeit focused on moral patiency rather than moral agency.

That said, I would favor giving implicitly or explicitly ethical care robots the ability to mimic human feelings. This particular type of mimicking is not deception. It is backed up by either implicit or explicit ethical principles genuinely followed by the robot. Compare the obviously blameable cases where humans hide intentions to harm while pretending to care, by faking friendly feelings.

*Alan Winfield*
Consider a simple thought experiment. You are walking on the street and notice a child who is not looking where she is going (perhaps she is engrossed in her smart phone); you see that she is in imminent danger of walking into a large hole in the pavement. Suppose you act to prevent her falling into the hole. Most bystanders would regard your action as that of a *good* person – in more extreme circumstances you might be lauded a hero. Of course, your action does keep the child safe, but it is *also* a moral act. Your behavior is consequentialist ethics in action because you have (a) anticipated the consequences of her inattention and (b) acted to prevent a calamity. Your act is ethical because it is an expression of care for another human's safety and well-being. To claim that AMAs merely exhibit safety+ behavior is to miss the key point that it is the *intention behind an act* that makes it ethical. Of course, robots do not have intentions – even explicitly ethical robots – but their designers do, and an AMA is an instantiation of those good intentions.

**Complexity**

*Susan Leigh Anderson & Michael Anderson*
The concern raised here by van Wynsberghe and Robbins (2018) has to do with the unpredictability of machine behavior in novel situations, making it difficult to ensure that it will behave ethically in these situations. With our approach to machine ethics, *all* the behavior of the machine is driven by a learned ethical principle that ideally can ensure that the machine will behave ethically even in unanticipated situations. This has been observed in our projects (Anderson and Anderson 2014; Anderson et al. 2017). Surely this is better than the track record of human behavior where there is not just an understated problem of unpredictability, but consistently unethical behavior by some is exhibited as well.

*Ben Byford*
The view that machines should be used only in places where they will do us no harm is again amicable. However, autonomous robots and algorithms are being developed around the world to automate everything automate-able. This is a thorny issue and concerns human value, purpose, and economic security. This has also fueled the conversation around universal basic income. Whether or not you believe this endeavor is ethically permissible, one can only imagine the evolution of Google's AlphaGo (as simply one example) will ultimately be used in more radical contexts than the 19x19 Go board to satisfy shareholders and human intrigue alike.

*Erica L. Neely*
I do not believe that we could simply rely on restricting the domain of interaction for a machine in the same way as AlphaGo, especially if we develop a truly autonomous machine or AI. These machines would likely experience a range of situations and would need the flexibility to respond to those situations – a capacity that AlphaGo does not require. Thus, while the desire to side-step the complexity problem is understandable, I am not convinced we can do so in the long run.

However, van Wynsberghe and Robbins' (2018) argument about unpredictable humans is problematic. It is true that society does place restrictions on humans who have seriously violated our norms. We do not accomplish this by restricting all humans, though – we restrict those who have violated the norms. We do not assume that simply because some humans behave badly that therefore all humans will behave badly. They state that "most of us assume that a random person will not intentionally cause us harm" (van Wynsberghe and Robbins 2018) – why assume that a random machine or AI would intentionally cause us harm either? There seems to be an odd double-standard where we are permitted to assume that humans mostly act ethically (with a few outliers) but that complex machines would not do so if they were AMAs. This seems unfounded.

*Alejandro Rosas*
As the example of AlphaGo shows, the issue of complexity cannot be reduced to unpredictability: AlphaGo, despite its unpredictability, needs no autonomous ethical capabilities. The issue of robot complexity concerns the scope of their capabilities to reason about all sorts of facts and to function autonomously in any domain of facts. AlphaGo's factual knowledge is restricted to the game of Go and its rules. If a freely ranging artificial agent can learn facts from any domain available to – or even beyond – humans in an interconnected coherent way, this is the same as having knowledge of the world at large. Such knowledge will include knowledge of social action and interaction, and with it of the potential harms to itself and to others as they arise from any source. All this knowledge will probably ensue from autonomous learning. It does not seem wise to create these machines without the further knowledge that harms are bad. Moreover, it would not make sense to block *ad hoc* their ability to learn the further rules (the moral rules) with which we humans deal with the 'bads' and 'goods' that arise from our own actions and those of others. For then they would just avoid harm to themselves with no regard to possible costs to others; a disastrous solution. The only way to altogether avoid going down the path that leads to autonomous artificial moral agents, would be to cut these super-knowers/learners from all capacity for action. This might not be technically feasible and attempting it could be even more hazardous than allowing them to learn and reason about morality.

*Alan Winfield*
All robots behave more or less unpredictably in the real world, simply because the real world (especially with humans in it) is itself unpredictable. It follows that I do not accept the argument that complexity is a special justification for AMAs. However, I do think that if AMAs ever became a practical reality they would need to be subject to new and tougher standards of verification and validation before being certified as safe for use, than non-AMAs, and a greater level of regulatory oversight when in operation.

**Public Trust**

*Adam Poulsen*
Machine ethicists argue that AMAs will increase the public's trust in the actions of robots. That argument here is that people will have more trust in a robot that can make explicit ethical decisions over one that does not. The most important thing to understand here is that AMAs are not commercial products, the increase in public trust machine ethicists are referring to is that when AMAs leave the research labs and enter the public sphere. To follow are several direct responses to van Wynsberghe and Robbins' (2018) article regarding the topic of public trust.

1. The geotagging example presented by van Wynsberghe and Robbins (2018) assumes that the technology behind AMAs will be used for ill purposes. This assumption regarding any emerging technology is not new, there are always concerns about technological advancement because of potential ill use. The possibility of intentional misuse is not a good reason to stop research into AMAs. AMAs are still restricted to research labs and no 'runaway' superintelligence has been developed, this research is still in its early stages. Moreover, AMA research could bring on a breakthrough in understanding ethics, computational reasoning, cognition, and agency. Lastly, such technology could largely better humanity, or at least the HRI experience.
2. I agree with van Wynsberghe and Robbins (2018) when they say that codes of conduct, standards, and certifications for AMA researchers need to be developed.
3. I agree when the authors state that "one might trust persons who are at the same time unpredictable" (van Wynsberghe and Robbins 2018). I have found this phenomenon in research concerning elderly care robots (Poulsen and Burmeister 2019). Acknowledging this phenomenon, AMA research should investigate what the social determinants of trust are and work them into the system, but not in a way that is dishonest, deceptive, or misrepresenting the ethics of a machine (Poulsen et al. forthcoming). My position argued in Poulsen et al. (forthcoming), is that trust in machines is developed by having the system's ethics demonstrable and designing the system to be sociable to evoke social determinants of trust.

*Susan Leigh Anderson & Michael Anderson*
We believe that trusting the behavior of a machine that affects humans *is* important and that transparency and consistency are critical to the public having that trust. Being *transparent* is why we insist on there being an ethical *principle* that drives autonomously functioning machine behavior which can be produced to justify its behavior (we do not rely on "algorithms [that] are black-boxed"), and clear cases of ethical dilemmas (that is, ones where the ethically relevant features and decisions are agreed upon by a consensus of applied ethicists) that led to clauses in the principle that justifies its actions can be examined. We aim to achieve this transparency in our work We also believe that *consistency* is important for trust and our approach to machine ethics does not permit inconsistency. Contradictions must be resolved in the learning process leading to the ethical principle that drives all of machine behavior.

### Ben Byford

Van Wynsberghe and Robbins (2018) state the following, "[w]ikipedia is an example of this form of trust as it requires trust not in persons but in the algorithms regulating the content on the website". This example is simply not the case unless I am getting confused at what level we are talking about the regulation coming into play. Wikipedia is regulated by participates of the platform and thus by humans (Wikipedia:About 2018). Humans write, edit and approve the contents of Wikipedia. Perhaps some better examples might be the recommendations system of Amazon.com (About Recommendations n.d.), searches using Google.com's search engine (How Search works n.d.), or the news feeds on Facebook (Constine 2016) which are solely algorithmic.

Trust is a tricky issue in technology as you must first trust that a technology works and that the intentions behind it are positive. Trust may indeed be misplaced in algorithms a lot of the time, but it is often not the algorithm itself that is leading to the loss of trust but the intentions behind the algorithms itself that we are losing trust in as they do something that misleads us. This can poison the well and make us mistrust the 'machine', or algorithm, as it were. Thus, it is important for designers and stakeholders to be transparent about the intention and capabilities of any technology, this is only becoming more important as we defer more judgement to autonomous systems. There is some great work being produced by (Theodorou et al. 2017; Wortham et al. 2016, 2017a, 2017b) investigating transparency of capabilities and decision-making within robotic systems.

### Erica L. Neely

When we trust a person, who or what are we trusting and why? We could say that we are trusting their moral code, because that is presumably what guides their actions. We could say that we are trusting their parents or how they were raised, because presumably those formative experiences shape who they are now. To some extent, probably all of these things go into trusting a person, because we recognize that they all contribute to a person's sense of morality. Children are shaped by their parents, by what they are taught and how they are raised, and ultimately by what moral code they adopt.

The same is likely true for AIs and advanced AMAs. To decide whether to trust them we will need to look at who created them and why, what they were programmed to do, and what kind of learning or alteration of that code is possible for them to do. I actually agree with the calls for transparency and standards for researchers who create such machines. I do not necessarily think that all contradictions can be resolved, as every ethical theory has its limits; if there were a single settled ethical theory, ethicists would not still have lively debates over what theory to follow. Yet, it is not at all clear to me that a theory needs to be perfectly consistent in order to be trustworthy. We trust other humans who are not transparent and who are not perfectly consistent in what they do; we trust them because of what they do most of the time. I suspect that is what we should be aiming for with AMAs. Do I think that there will never be a situation in which an AMA makes a mistake or in which an AMA does not know what to do? No. But that is true of humans as well. People who demand perfection from machines will likely never trust them. However, people with more reasonable expectations should be satisfied by the calls for transparency made by machine ethicists.

*Alejandro Rosas*
van Wynsberghe and Robbins (2018) claim that there is "inconsistency between the promotion of AMAs for reasons of complexity and for reasons of trust: it is inconsistent to expect unpredictability in a machine and to expect trust in a machine at the same time". I believe they misunderstand the connection between complexity, unpredictability and trust. Morally risky unpredictability arises from complexity only when robots have reached the point where they are able to ascertain the facts of a situation in any arbitrary domain of facts, *while lacking at the same time* moral decision-making capabilities. As I pointed out previously, AlphaGo is "unpredictable" but not in this sense. The risky type of unpredictability arises if free ranging robots understand the facts of the world in any domain while altogether lacking moral directives; and this is surely a reason to promote decision-making capabilities in those robots, i.e., converting them to AMAs. This is perfectly consistent with making them capable of earning the trust of humans. It could be argued that if they can autonomously learn all types of facts, they could also autonomously learn moral rules. It will be prudent, nonetheless, to guide their moral learning, as we do with our children, especially considering that moral norms are subject to disagreement and social debate. And it might be a good idea to hard-wire some extra constraints, e.g. regarding self-conceited behavior. Humans are biologically hardwired for both egocentric bias and impartial morality. AMAs should only be hardwired for the latter.

*Alan Winfield*
I am also skeptical of the argument that AMAs will increase public trust. And of course, until there are field trials that properly test such an assertion, we simply do not know whether the argument is justified or not. I agree with my co-respondents here that transparency is of key importance – for *all* robots whether AMAs or not. I would go further and argue that we need *explainability*. An elderly person should, for instance, be able to ask her care robot 'why did you just do that?' and receive a straightforward plain language reply. If that robot were an AMA it should also be able to answer questions like 'what would you do if I fell down?' or 'what would you do if I forget to take my medicine?' These explanations would help her to build a predictive model of how the robot will behave in different situations. If the robot consistently behaves as she expects then, over time, she may come to trust it, but I am doubtful that trust would be correlated with the robots 'ethical' behaviors.

**Preventing Immoral Use**

*Adam Poulsen*
Concerning the prevention of immoral use as a reason to develop AMAs, van Wynsberghe and Robbins (2018) discuss the issues of attempting to program an AMA breathalyzer with deontologist or utilitarian ethical governance. In the author's example, an intoxicated woman attempts to escape domestic violence in a car but it will either never start the engine because one should never drive while intoxicated (if it has deontologist principles) or it might start to save the woman's life (utilitarian) (van Wynsberghe and Robbins 2018). However, as the authors note, there is an issue with the utilitarian breathalyzer that it might allow the driver to kill two other people if it starts the car (van Wynsberghe and Robbins 2018). There are some points to be made here, they are as follows.

1. My position is that normative ethical theories, such as deontology and utilitarianism, are not the best way to program machine morality, they are not even the best way to wholly represent human morality. Some machine ethicists favor providing an AMA with a set of normative ethical principles, this is called the top-down approach (Wallach et al. 2008). However, many prefer the bottom-up approach with machine learning or case-based learning (Shaw et al. 2018), the hybrid approach with elements of both top-down and bottom-up approaches (Charisi et al. 2017), the value-based approach (Poulsen and Burmeister 2019), the human-like moral competence approach (Torrance and Chrisley 2015), and so on. Evidently, there are multiple ways of endowing machines with morality, and some are better at representing human reality than others.
2. I would never suggest that a breathalyzer be an AMA because such a machine's central goal is to ensure the safety of all stakeholders. As stated previously, my criteria for a machine being permitted to be an AMA is that users interacting with it must not experience physical or psychological harm because of the AMA making dynamic ethical decisions which overlap safety measures or its core goals. A breathalyzer being an AMA could be dangerous because it may encounter a dilemma, such as that described by the authors, in which it must make a potentially harmful decision either way. Not all machines should be AMAs, intrinsic safety through implicitly ethical behaviors and design should be the priority.

The author's offer another example, wondering if a care robot will fetch a patient a glass of wine that would give them immediate pleasure but might affect their long-term wellbeing (van Wynsberghe and Robbins 2018). This example is concerned with the weighing of consequences and understanding of context (the frame problem), as performed by the care robot. My position in favor of these types of trade-offs is established in my work on the design of elderly care robots with Values in Motion Design in Poulsen and Burmeister (2019) and Poulsen et al. (2018a). Intrinsic care values, such as wellbeing, should always be ensured (by design) over extrinsic care values such as pleasure.

*Susan Leigh Anderson & Michael Anderson*
The ethical dilemmas introduced by the authors in this section confirm the approach that we have taken to understanding and representing ethical dilemmas, as well as resolving them. If there is a true ethical *dilemma*, then it is the case that two or more *prima facie* duties pull in different directions: e.g. in one of the authors' examples, respect for the autonomy of the patient would lead the care robot to think that it should accede to the request to give the person a fourth glass of wine, whereas the duty to prevent harm might lead to rejecting the request on the grounds that acceding to it could be harmful to the patient. These are the sorts of dilemma cases that should be resolved by the learned ethical principle. Input from the patient's doctor would undoubtedly be an important factor for an elder care robot designed to function with this patient. If the doctor has instructed the robot to contact her if the patient starts to drink heavily (more than a certain amount of alcoholic drinks in so many hours) because it would be a sign that he is becoming depressed or it would dangerously conflict with medications he is taking, then the care robot will know that in this case it should contact the doctor and not accede to the patient's request.

## Ben Byford

With van Wynsberghe and Robbins' (2018) arguments how "should the device be programmed?" and "if one is claiming that robots should be involved in the decision-making procedure it must be very clear how a 'good' robot is distinguished from a 'bad' one", we are getting to beating heart of machine ethics. My position is that it is important that we have well considered research in this territory so that when we are faced with practical reality of programming said technologies we can do so with the best intentions for positive outcomes.

To give some practical answers, capability decision trees in conjunction with machine-learned, implicit rules can begin to satisfy a simple robotic system operation giving transparency through the knowledge of what it is capable of doing, simulations of past behavior, and hard coded rules it must follow. As one example Anderson et al. (2017) have produced work using some of these techniques.

In my opinion, van Wynsberghe and Robbins' (2018) statement concerning the "good" and "bad" is subtler and possibly misplaced. Firstly, transparency will go some way to help us understand what the capabilities are, as well as a computed decision from a given AMA. Personally, I am unable to attribute personified terminologies such as "good" and "bad" to machine systems but find no issue with their creators. The endeavor of machine ethics should eventually help industries create products and services that have the best intensions, hold categorical ethical knowledge, and allow for ethical decision-making in line with human wellbeing - all of which should be discussed openly as a public with support and interrogation from academia.

## Erica L. Neely

I think this is a pretty bad reason for developing AMAs. Almost any object can be misused; I can drop a heavy object out of a window on to someone's head and harm them, but we do not try to ensure all objects are light as a result. Requiring robots or machines to be free from misuse seems unreasonable. With that said, I believe that part of the difficulty van Wynsberghe and Robbins (2018) raise comes from a lack of clarity about the purpose of the machines in question. For instance, they present the case where an elderly person asks a robot to fetch him or her a fourth glass of wine and ask whether the robot should do so. The answer to this largely depends on the purpose of the robot.

Aristotle (2000) saw goodness as a matter of something performing its function well. Goodness for a thing therefore depends on its function – I cannot tell you whether this is a good knife without knowing whether you want to use the knife to butter bread or to butcher a deer. The same knife will not be good at both of those tasks. Similarly, we need to know the function of the robot. If the robot is a care robot, tasked with looking after the health of the elderly person, then it should likely refuse to bring the wine. Assuming that the person has consented to the robot's presence, he or she has chosen to allow for somewhat paternalistic behavior in the service of improved health. While the person's autonomy is being infringed upon in this specific case, the person has implicitly given prior consent to this infringement. Of course, if the care robot is being deployed without the consent of the person in question, there will be many ethical issues raised about autonomy, paternalism, privacy, and so forth, and the fact that a machine is involved will be fairly secondary to those larger concerns.

On the other hand, if the robot exists simply for utility purposes – to be a second set of hands or be more mobile than its owner – then it should bring the wine, because its purpose is simply to extend the function of its human owner. We make extendable grabbers for people who need to reach something off of a high shelf or off of the floor and who cannot reach those places for themselves. This could be due to a physical impairment such as having lost the flexibility to bend over due to spinal fusion or even just because a person is too short to reach high shelves. The grabber is simply an extension of its user; it does not evaluate whether it is trying to reach a syringe of heroin or a sock. Moreover, it seems unlikely that we actually would like the grabber to engage in moral debate over whether to grab the thing we are trying to reach; it is simply filling in for something we cannot (or can no longer) do physically ourselves. Similarly, a utility robot that functions in this fashion may well not need to be an AMA, because that is not why we have developed it; its function does not demand it.

A care robot could well be improved by being an AMA, since it would more easily be able to deliberate about the wine case. I made a prima facie case for the robot's refusal based on the patient's prior consent above, hinging on the fact that a fourth glass of wine is likely to have a detrimental effect on the person and the person has tasked the robot with looking after his or her health. However, this glosses over the fact that even in healthcare settings we acknowledge the right to refuse or withdraw from treatment; it is not the case that we can always act paternalistically towards patients. Furthermore, we cannot simply reduce healthcare to health improvement. If the patient were in a palliative care setting, where the job of a health care professional is simply to make them comfortable until they died, then the fact that a fourth glass of wine is detrimental to their health would likely not matter so long as it would be beneficial to their overall mental wellbeing. Since care is such a complicated issue, it may well be that such a robot should be an AMA since it could consider all of these factors (rather than simply assuming that fetching the wine would be immoral, as seems to be supposed by the title of this section.) The robot's function is a key factor in whether it should be an AMA and in whether it is being used for "good" or for "bad."

*Alan Winfield*
AMAs *themselves* pose a risk of immoral use either because the robot is 'gamed' (i.e. tricked into misbehaving) or, more seriously, through accidental or malicious hacking. In recent work we demonstrated that an ethical robot can be transformed into a competitive or aggressive robot through the trivial alteration of less than one line of code (Vanderelst and Winfield 2018b). The ease of this transformation is a consequence of the fact that the robot's ethical rules are expressed in a few lines of code; that code then becomes a single point of vulnerability. Of course, any robot can be hacked, but we argue that AMAs are unusually vulnerable because the potential impact of their ethical rules being compromised is so dangerous.

**Morality: Better with Moral Machines & Better Understanding of Morality**

*Adam Poulsen*
In their critique, van Wynsberghe and Robbins (2018) argue that by providing morals to AMAs it might cause humans to become morally deskilled (they make this statement with supporting literature in which machine ethicists have claimed that AMAs will become more moral than humans). This argument is like when Socrates said that writing would cause forgetfulness: "this invention [writing] will produce forgetfulness in the minds of those who learn to use it, because they will not practice their memory" (Plato 1925). Socrates' claim has never been shown to be true and I would expect the same result from the claim that giving morals to machines will morally deskill humans. Furthermore, the moral deskilling of humans presumably relies on people not interacting morally with the machines and with other humans or non-human agents. As long as humans exist, there will be people morally interacting with each other. Thus, there is no point at which humans will not require their moral reasoning unless they do not interact with anyone, or anything, else.

As the authors point out, programming morality requires "an understanding of moral epistemology such that one can program machines to "learn" the correct moral truths - or at least know enough to have AMAs learn something that works" (van Wynsberghe and Robbins 2018). It is undeniably true that this is an issue that creates a barrier to programming morals. This issue needs to be discussed by great minds in philosophy and ethics, and therefore it will undoubtedly advance our understanding of morality in the process. The existence of this issue is why more research needs to be done, not why it needs to stop.

The author's raise the issue of unknown moral truths being a barrier to AMA development – I agree. In my opinion, until absolute moral truths or 'goods' are known we should provide AMAs with limited moral freedom. To do this, we program in what we currently consider to be certain moral truths and limit the AI's freedom to guarantee that safety concerns are respected first. Then, within a fixed context, let the AMA customize its behavior to respect those moral truths in a personalized way for each individual, only when it is sensible to do so. In this way, moral truths are always respected, but the importance of them and how the robot behaves to respect them depends on the user's preferences. This differs from letting an AMA freely 'do ethics' because of the limitations placed on the machine. The actions are not globally applied in a standardized way, instead, they are customized to each person. It is as if the person was programming the robot themselves within specific safety limits.

### *Susan Leigh Anderson & Michael Anderson*

We agree with Dietrich (2001) that, in general, humans are far from ideal ethical reasoners, and one explanation for this lies in the fact that we have developed evolutionally to favor ourselves and our group, which often leads to unethical behavior. Since machines can be designed without this evolutionary bias, they could be taught to be more ethical. We think we have a valid approach to representing ethical dilemmas, teasing out ethically relevant features and corresponding prima facie duties, leading to an ethical decision principle for a machine that functions in a particular domain, from applied ethicists' examples of cases where the correct action is clear. Our position and work related to this can be found elsewhere (Anderson and Anderson 2014; Anderson et al. 2017). The process of doing so will very likely bear fruit in the study of ethics: (1) by discovering principles implicit in the considered judgements of ethicists that have not been stated before, (2) by forcing the examination of inconsistencies that need to be resolved, and (3) by thinking about real life dilemmas rather than the extreme hypothetical examples discussed in ethical theory courses. (4) We, also have a chance of considering ethics from a fresh perspective: how we believe machines should treat us, rather than how we can maintain/improve our personal positions in life. We have said that the new perspective will be similar to Rawls' thought experiment of coming up with the principles of justice behind "a veil of ignorance", of not knowing who you will be in society, so you will consider that you might be the most needy/disadvantaged person.

We believe it is unfair to reject the entire field of machine ethics on the grounds that there are certain extreme ethical dilemmas about which there is no agreement at this time (at least by laypersons and in some cases even ethicists) as to how a machine ought to behave. We argue that where there is not a consensus as to ethically correct behavior that a machine functioning autonomously in a particular domain might face, that we should not allow machines to function autonomously in this domain.

One reason, we believe, why many have adopted the position that if it does not seem possible to resolve ***all*** ethical dilemmas that some proposed machine might face if allowed to function autonomously, then we should completely reject the validity of the field of machine ethics, is because some think that every ethical dilemma involves life or death decisions, and of course these are contentious. But, not all ethical dilemmas involve life or death. The elder care assistant robot that we have been developing (see Anderson et al. (2017)) must weigh the strength of many prima facie duties (e.g. respecting the autonomy of the patient, being faithful to a doctor's instructions [including medication reminding], avoiding harm to the patient, looking for signs that the patient is incapacitated, offering to do tasks for the patient, making sure that it is charged enough to perform its duties) continually, and these are all ethical dilemmas. And it is important to be reminded of what we said earlier, that there are machines in existence today or on the immediate horizon that will function autonomously without ethical constraints unless we insist that ethical principles are instilled in them.

We would like to defend a position that lies between Dietrich's (2001) and the authors' portrayal of Aristotle's. We believe that by developing machines (e.g. robots) that behave more ethically than most humans, they can serve as good role models for us, interaction with which might lead to an improvement in our own behavior. Surely this would be better than allowing us to continue to do terrible things to our fellow human beings (offenses which are not, after all, victimless) through our continued "practice" of slowly "gaining moral understanding" when help in speeding up this process might be possible.

### *Ben Byford*
"The point is that human morality… is dependent upon many complex factors and building a machine that tries to perfectly emulate human morality must use each of these factors combined rather than rely on ethical theory alone" (van Wynsberghe and Robbins 2018). Here, the authors argue against AMA research as morality is both too complex and embodied. If this is the general consensus than the field may indeed be in trouble. However, we could also endeavor to create a moral subset, a new morality or discover a truer morality while endeavoring with AMAs. The journey to finding a morality that is both computable and permissible, subjected to humans, may ultimately fall flat, but the endeavor is not limited to the above creation of a computational moral nirvana, it is focused on creating knowledge that can both expand our moral understanding in even a small way while discovering the limits of computable moral reasoning for practical implementation.

### *Fabio Fossa*
In my opinion, van Wynsberghe and Robbins adopt a too narrow and one-sided conception of AMAs. In more than one occasion, in fact, they seem to claim that AMAs, as "machines… endowed with moral reasoning capabilities" (van Wynsberghe and Robbins 2018), must be conceived of as homogeneous to human moral agents (HMA), i.e., as (at least to-be) full moral agents. This, however, is just one of the possible conceptions of AMAs–and, I think, not even the most accurate one.
As the authors suggest, the notion of AMA belongs first and foremost to machine ethics. Machine ethics is a subfield of research in engineering, mainly involving computer science, AI and robotics. Its purpose, one might say, is to design and build AMAs. In order to understand what is precisely meant by AMAs, therefore, it becomes pivotal to clarify the way in which the notions of "agency" and "morality" are defined in this specific epistemological domain, so as to distinguish them from the way in which they are commonly used.

In computer science, robotics and AI an agent can be defined as "a system situated within and a part of an environment that senses that environment and acts on it, over time, in pursuit of its own agenda and so as to effect what it senses in the future" (Franklin and Graesser 1997) – where the idea of "an agenda of its own" must not be overrated (Jonas 2001), since it includes the execution of tasks whose ends are already given.

Although this definition may seem too broad from a philosophical perspective (it applies both to human beings and thermostats), it serves the purpose of the technological project in the context of which it is used, that is, to build artificial agents. A similar point can be made regarding the main features of such technical notion of agency: autonomy, interactivity, adaptability (Floridi and Sanders 2004), all described in quantitative terms and according to the logic of efficient causality so that they can be phrased in the language of technological science–or, in other words, so that they can be understood at the *Level of Abstraction* of computer science, robotics and AI.

To the same group belongs the feature of "moral agency", defined as the capability of performing "morally qualifiable action(s)" where "an action is said to be morally qualifiable if and only if it can cause moral good and evil" (Floridi and Sanders 2004) – regardless of who actually decides what should be taken as morally good or evil. In this context, "moral reasoning" would then be the ability to autonomously select between alternative scenarios in accordance with constraints that may be generally considered as displaying moral meaning. This technical conception of moral agency, which underlies the notion of AMA, is entirely functional to the aim of machine ethics: in fact, it allows envisioning "moral machines" and elaborating an hypothetical framework for their design. Hence, it makes sense primarily within the epistemological domain of machine ethics.

When the relation between AMAs and human moral agents (HMAs) is tackled, however, the debate goes beyond the epistemological domain of machine ethics and enter the domain of ethics (or, more precisely, machine metaethics). The problem, now, is how to translate the technical concept of AMA in philosophical language. What might the notions of artificial agency and morality mean in this different dimension?

From this perspective, AMAs have been mainly understood either as homogeneous to HMAs, and therefore bound to replace us and make us obsolete, or as "morally sensitive" technological products, i.e., tools that execute functions autonomously (that is, without requiring constant human intervention or supervision) in accordance to given constraints that carry a moral weight for the people involved in their design and deployment. Overlooking this distinction, the authors' criticism–one that I agree with on many points, especially the last ones–applies exclusively to those who believe that AMAs and HMAs are essentially equal and only incidentally different from one another. However, this is neither the only way to translate the technical notion of AMA in philosophical language nor, in my opinion, the most accurate way to do this. Once the alternative conception of AMAs as morally sensitive autonomous executers of functions is endorsed, most of the authors' claims fail to convince. This last conception is in my opinion the best way to think about AMAs from a philosophical point of view (Fossa 2018) and to address the many ethical problems they pose.

Finally, I do not agree that scholars as Johnson, Bryson and Tonkens (I am not sure about Sharkey) would be against the development of this kind of AMAs, as van Wynsberghe and Robbins (2018) write. On the contrary, I believe that they criticize the notion, the development, or the feasibility of AMAs as responsible, full moral agents and that, in doing this, they lay down the theoretical principles to interpret AMAs as morally sensitive autonomous tools–the morality of which lies entirely in our eyes and hands.

### Erica L. Neely

It is unlikely that we will outsource all of our moral decision-making, in the same way that I do not outsource all of my arithmetic to a calculator. Sometimes you do not wish to consult a machine; you wish to do something yourself. So, the worry that we will become less skilled at moral decision-making seems unlikely, especially as it seems to presuppose we will use AMAs in all of these cases.

It is unreasonable to require that AMAs wait until we have established a set of principles as definitely correct, since we do not do that with humans. We are in ethical situations now and we have to do something; we cannot simply avoid them while hoping ethicists will eventually provide us with ultimate ethical truths. Similarly, if we place machines in situations where decision-making of this sort needs to happen (or if they are AIs, they may place themselves in that situation), then they are going to have to do the best they can, just like the rest of us.

I do not think that AMAs will be better or worse than us at moral thinking, other than perhaps in the same sense that an ethicist is better at thinking through a moral decision than a lay person due to experience with nuanced thought. A machine might be better at keeping all of the factors in mind, so to speak, but beyond that it will be in the same situation as any of the rest of us – trying to use its principles to make the best decision it can, without any guarantee that it is correct. What I think AMAs truly do is force us to confront ethical dilemmas we have mostly ignored. The problems with autonomous vehicles, for instance, could be faced by *any* driver – anyone could be in a situation where they have an unavoidable collision and have to decide what to run into. We just mostly avoid thinking about that situation. Similarly, I believe that AMAs throw ethical problems into stark relief by forcing us to think about ethical dilemmas that we prefer to ignore. They are not making new problems; they are forcing us to confront problems that already exist.

### Alejandro Rosas

There is a lot to be said in favor of the view that AMAs will be morally superior to humans. Humans have been endowed by natural selection with special psychological mechanisms that favor cooperative behavior, precisely because our natural behavior without these special mechanisms favors a selfish drive to put oneself above others. This selfish drive alone leads to strictly linear dominance hierarchies. It favors neither the large-scale quasi-egalitarian cooperation that we see in hunters and gatherers nor the admittedly less egalitarian but even larger in scale cooperation that we see in agricultural and industrial societies. We can thus say that humans are designed to be selfish and moral at the same time and show clear individual and collective signs of this unpromising dual nature. The frailty of human morality, even when manifested in reasoning failures, is not due to lack of knowledge of moral norms, but rather to the constant – though individually variable – temptation to self-conceit, nepotism and ingroup bias. With AMAs we have the possibility of creating a moral entity that is free from these frailties, a "holy will" in Kantian terms, meaning that its behavior is necessarily and spontaneously aligned with the moral law. Intentionality and freedom, in the compatibilist sense, will be reached by AI as soon as it reaches human level intelligence. To insist that we should not attempt to create AMAs as long as we lack standards of moral truth, brings in philosophical skepticism where it does not belong. Since I doubt that many of us would be willing to skeptically sweep aside the declaration of human rights and our democratic legal systems as a fiction, we already have a consensus, a minimal body of knowledge that we know AMAs should share with us. We certainly face problems to determine the right thing to do when rights conflict; but even when facing such conflicts, we are far from arguing in the dark. Compromises are feasible. The search for absolute moral truths is meta-ethically suspect (Mackie 1977). Pragmatically, it is not likely to bring us nearer to consensus, but rather to breed intolerance and social paralysis.

### *Alan Winfield*

In response to the point 'morality: better with moral machines', I think we need to make the distinction between what might in principle be achievable in the near future and the far future. For machines to be as good as humans at moral reasoning they would need, to use Moor's terminology, to be *full ethical agents* (Moor 2006). The best we have demonstrated to date is a handful of proof-of-concept explicitly ethical agents and even those, as I have commented, are only minimally ethical. In no sense are such agents better at moral reasoning than humans.

Does the fact that we have arguably reached the third category in Moor's scheme (explicit ethical agents) mean that full ethical agents are on the horizon? The answer must be no. The scale of Moor's scheme is not linear. It is a relatively small step from ethical impact agents to implicit ethical agents, then a very much bigger and more difficult step to explicitly ethical agents, which we are only just beginning to take. But then there is a huge gulf to full ethical agents, since they would almost certainly need something approaching human equivalent AI (full explicit ethical agents, by Moor's (2006) definition, require consciousness, intentionality, and free will). Indeed, I think it is appropriate to think of our explicitly ethical machines as examples of narrow AI, whereas full ethical agents would almost certainly require artificial general intelligence (AGI).

Much of the argument in van Wynsberghe and Robbins (2018) relating to this question actually refers to full ethical agents, which are – as far as I am concerned – such a distant prospect as to be of theoretical interest only, and not a practical concern. And I am not persuaded that even full ethical agents would be better moral reasoners than humans.

On the final point 'better understanding of morality' here I would argue that building computational models can and is providing insights that cannot easily be obtained with human subjects. As Richard Feynman famously said, "what I cannot create, I do not understand". In Winfield et al. (2014) we demonstrated for the first time a minimally ethical robot facing a balanced ethical dilemma. One of several insights from these experiments is that the idea that a robot can resolve ethical dilemmas better than humans is a fallacy. This, and more recent work, provides not just a demonstration of simple ethical reasoning but more generally tests an embodied computational model of the simulation theory of cognition (Vanderelst and Winfield 2018a). These are of course very simple experiments, which are far from providing an understanding of human morality, but as in all things we have to start somewhere.

## Conclusion

The purpose of this paper is to both defend and critique the reasons for developing AMAs while responding to van Wynsberghe and Robbins' (2018) paper *Critiquing the Reasons for Making Artificial Moral Agents*. In doing so, this paper adds, we hope, to the robot ethics and machine ethics conversation by featuring the unique research approaches and goals, as well as ultimate intentions for AMA uses, of a number of machine ethics researchers. Also contributing here are well-established commenters on machine ethics who come from philosophy and the public sphere. Not every contributor here agrees with the reasons posited by some machine ethicists in literature for pursing AMA research, some strongly disagree. However, all authors do agree that AMA research is useful and worthwhile. Moreover, we agree that a strong ethical approach to developing ethical machines is needed – the approach favored by van Wynsberghe and Robbins (2018).

Responses to a Critique of Artificial Moral Agentsaccounts for the values of those affected by its use. *Ethics and Information Technology, 18*(3), 185-198, doi:10.1007/s10676-016-9404-2.

Burmeister, O. K., Bernoth, M., Dietsch, E., & Cleary, M. (2016). Enhancing connectedness through peer training for community-dwelling older people: A person centred approach. *Issues in Mental Health Nursing*, 1-6. doi:10.3109/01612840.2016.1142623

Burmeister, O. K., & Kreps, D. (2018). Power influences upon technology design for age-related cognitive decline using the VSD framework. *Ethics and Information Technology, 20*(3), 1-4. doi:10.1007/s10676-018-9460-x

Charisi, V., Dennis, L. A., Fisher, M., Lieck, R., Matthias, A., Slavkovik, M., et al. (2017). Towards Moral Autonomous Systems. *CoRR, abs/1703.04741*.

Constine, J. (2016). How Facebook News Feed Works. TechCrunch. https://techcrunch.com/2016/09/06/ultimate-guide-to-the-news-feed/?guccounter=1.

Dietrich, E. (2001). Homo sapiens 2.0: why we should build the better robots of our nature. *Journal of Experimental & Theoretical Artificial Intelligence, 13*(4), 323-328, doi:10.1080/09528130110100289.

Draper, H., & Sorell, T. (2017). Ethical values and social care robots for older people: an international qualitative study. *Ethics and Information Technology, 19*(1), 49-68, doi:10.1007/s10676-016-9413-1.

Floridi, L., & Sanders, J.W. (2004). On the Morality of Artificial Agents. *Minds and Machine, 14*, 349-379, doi:10.1023/B:MIND.0000035461.63578.9d

Fossa, F. (2017). Creativity and the Machine. How Technology Reshapes Language. *Odradek. Studies in Philosophy of Literature, Aesthetics and New Media Theories, 3*(1-2), 177-213.

Fossa, F. (2018). Artificial Moral Agents: Moral Mentors or Sensible Tools? *Ethics and Information Technology, 2*, 1-12, doi:10.1007/s10676-018-9451-y.

Franklin, S., & Graesser, A. (1996). Is it an agent, or just a program? A taxonomy for autonomous agents. In J.P. Müller, M.J. Wooldridge, & N.R. Jennings (Eds.), *Intelligent Agents III. Agent Theories, Architectures, and Languages. ATAL 1996. Lecture Notes in Computer Science (Lecture Notes in Artificial Intelligence),* vol. 1193 (pp. 22-35). Berlin, Heidelberg: Springer

Garner, T. A., Powell, W. A., & Carr, V. (2016). Virtual carers for the elderly: A case study review of ethical responsibilities. *Digital Health, 2*, 1-14, doi:10.1177/2055207616681173.

Goodall, N. J. (2014). Ethical Decision Making during Automated Vehicle Crashes. *Transportation Research Record, 2424*(1), 58-65, doi:10.3141/2424-07.

How Search works (n.d.). Google. https://www.google.com/search/howsearchworks/.

IEEE Standards Association (2017). Ethically Aligned Design: A Vision for Prioritizing Human Well-being with Autonomous and Intelligent Systems.

International Organization for Standardization (2016). Robots and robotic devices -- Safety requirements for personal care robots. https://www.iso.org/standard/53820.html.

Jonas, H. (2001). A Critique of Cybernetics. In *The Phenomenon of Life: Toward a Philosophical Biology* (pp. 108-127). Evanston: Northwestern University Press.

Landau, R. (2013). Ambient intelligence for the elderly: Hope to age respectfully?
33